\documentclass[12pt]{extarticle}
\usepackage[utf8]{inputenc}
\usepackage[utf8]{inputenc} % allow utf-8 input
\usepackage[T1]{fontenc}    % use 8-bit T1 fonts
\usepackage{hyperref}       % hyperlinks
\usepackage{url}            % simple URL typesetting
\usepackage{booktabs}       % professional-quality tables
\usepackage{amsfonts}       % blackboard math symbols
\usepackage{nicefrac}       % compact symbols for 1/2, etc.
\usepackage{microtype}      % microtypography
\usepackage{amsmath}        % microtypography
\usepackage{lipsum}
\usepackage{color}
\usepackage{caption}
\usepackage{wrapfig}
\usepackage{enumitem}

\usepackage{cite}
\newtheorem{theorem}{Theorem}

\title{Proof of the impossibility of probabilistic induction}
\author{Vaden Masrani}
\date{September 12, 2019}

\begin{document}

\maketitle

In what follows I restate and simplify the proof of the impossibility of probabilistic induction given in \cite{popper1985,popper1992}. Other proofs are possible (cf.~\cite{popper1985}).

\section{Logical Entailment}
Given two statements $x, y$ we write $x \vdash y$ if $x$ \textit{entails} $y$.
For example, if
\begin{align*}
    &x~=~\textit{``All men are mortal''}\\
    &y~=~\textit{``Socrates is a man''}\\
    &z~=~\textit{``Socrates is mortal''}
\end{align*}
we can write $(x \wedge y) \vdash z$. Entailment is a ``truth broadcasting'' operation that allows the truth of the premises to flow to the conclusion.

We also have the following simple theorem.
\begin{theorem}
    If $x \vdash y$ then $p(y|x) = 1$.
\end{theorem}
For example if
\begin{align*}
    &x~=~\textit{``All swans are white''}\\
    &y~=~\textit{``All swans in Austria are white''}\\
\end{align*}
Then $x \vdash y$ and $p(y|x)=1$ reads ``The probability that all swans in Austria are white given that all swans are white is 1.''

\section{The Problem of Induction}

We describe the classical problem of induction by first describing (the non-problem of) deduction.
We have a ``general law'' statement $G$ which is assumed to hold across time and space, and we have a set of observed evidence statements $e_1, e_2, ..., e_N$ that are assumed to be true in a particular spatio-temporal region.
We refer to the conjunction of the observed evidence as $E = (e_1 \wedge e_2 \wedge ...)$.
By definition the general law $G$ logically entails each of the evidence statements
\begin{align}
    &G \vdash e_i
    &\forall e_i \in E
\end{align}
For example, consider evidence statements
\begin{flalign*}
    &e_1~=~\textit{``All observed swans in Austria in 1796 are white''} \\
    &e_2~=~\textit{``All observed swans in Austria in 1797 are white''} \\
    &...
\end{flalign*}
And consider the generalization
\begin{flalign*}
    &G=\textit{``All swans are white''}
\end{flalign*}

Then $G \vdash E$ is logically valid and referred to as ``deductive inference''. However, $E \vdash G$ is logically \textit{invalid}. This is the classical ``problem of induction''. It states that no amount of evidence statements $e_i$ can \textit{logically} justify a generalization $G$.

\section{Probabilistic Solution to the Problem of Induction}

The probabilistic solution states that while the truth of $G$ cannot be logically established, the \textit{probability} of any general law $G$ can be established and increases with the accumulation of favorable evidence.
Therefore, while we cannot say $G$ is true, we can say $G$ is \textit{probably} true.
The argument is as follows.

Let $G$ and $E$ be as above. Let $0 < P(G) < 1$ be our prior belief in $G$ and let $0 < P(E) < 1$ be the probability of the evidence. Because $G \vdash E$ we have $P(E | G) = 1$. Therefore using Bayes rule we arrive at the following:

\begin{align}
    &P(G | E) = \frac{P(E | G) P(G)}{P(E)} = \frac{P(G)}{P(E)} \nonumber \\ \intertext{And because 0 < P(E) < 1}
    &P(G | E) > P(G)
\end{align}

Therefore the probability of $G$ increases with favorable evidence $E$. This seemingly justifies a belief in probabilistic induction. In the next section it will be shown that despite this seeming plausibility, probabilistic induction is impossible.

\section{Proof of the Impossibility of the Probabilistic Solution to the Problem of Induction}

Again let $E = (e_1 \wedge e_2 \wedge ...)$ where
\begin{align*}
    &e_1~=~\textit{``All observed swans in Austria in 1796 are white''} \\
    &e_2~=~\textit{``All observed swans in Austria in 1797 are white''} \\
    &...
\end{align*}
Now instead consider two competing generalizations of the evidence $E$
\begin{align*}
    &G_g=\textit{``All swans are white''} \\
    &G_b=\textit{``All swans are violet except in Austria where they are white''}.
\end{align*}
$G_g$ is a typical ``good'' generalization while $G_b$ is a ``bad'', or ``anti-inductive'' generalization of the evidence. Being Bayesian, we can give the prior probabilities of $P(G_g)$ and $P(G_b)$ any values between 0 and 1 we like.

We will consider the two ratios
\begin{align*}
    &R_{\text{prior}} = \frac{P(G_g)}{P(G_b)}
    &R_{\text{posterior}} = \frac{P(G_g | E)}{P(G_b | E)}
\end{align*}

Because any generalization by definition entails the evidence, we have $P(E|G_g) = P(E|G_b) = 1$. Therefore we have the following theorem
\begin{theorem}
    \begin{equation}
        R_{\text{posterior}} = \frac{P(E|G_g)P(G_g)/P(E)}{P(E|G_b)P(G_b)/P(E)} = R_{\text{prior}}
    \end{equation}
\end{theorem}

This shows that inductive learning can never favor one generalization over another. Despite raising the probability according to (2), it raises the probability of \textit{all} generalizations, even anti-inductive ones such as $G_b$. Or, in Popper's words\cite{popper1992}:

\textit{``Theorem (2) is shattering.
It shows that the favourable evidence $E$, even though it raises the probability according to (2), nevertheless, against first impressions, leaves everything precisely as it was.
It can never favour $(G_g)$ rather than $(G_b)$.
On the contrary, the order which we attached to our hypotheses before the evidence remains.
It is unshakable by any favourable evidence.
The evidence cannot influence it.}''

Therefore probabilistic induction cannot favor inductive generalizations over anti-inductive generalizations, and we conclude that probabilistic induction is impossible.

\section{Commentary}

The proof relies on the fact that, for any generalization $G_i$ of the evidence $E$, the likelihood $P(E|G_i) = 1$. This is the formal condition for induction -- to induce general laws of nature from observations. If one wished to refute the proof, they would have to either claim:
\begin{enumerate}
    \item $G_i$ isn't a general law, which means probabilistic induction is not capable of inducing general laws, or
    \item $G_i$ is a general law but that $P(E|G_i)~\neq~1$. Stated using our example, this says: ``the probability that all swans in Austria are white is not 1, despite the fact that all swans are white.'' In other words, the general law $G_i$ is not a general law, which is a contradiction.
\end{enumerate}

This shows that probability calculus is not capable of discovering (i.e. inducing) general laws of nature from data. Given that human beings \textit{are} capable of discovering general laws of nature from data, this further shows that the products of human cognition are not products of the probability calculus.

\bibliographystyle{plain}
\bibliography{M335}

\end{document}